# Opto-Electronic Hybrid Reinforcement Learning Based on MZI for Robotic Manipulator Control

Mengting Yu[1], Shuiying Xiang[1*], Changjian Xie[1*], Yonghang Chen[1], Haowen Zhao[1], Xingxing Guo[1], Zhijian Hao[1], Yahui Zhang[1], Yanan Han[1], & Yue Hao[1]

[1] State Key Laboratory of Integrated Service Networks, State Key Discipline Laboratory of Wide Bandgap Semiconductor Technology, Xidian University, Xi'an 710071, China.

## Abstract

Precise motion control for autonomous robots is a core technical enabler for the large-scale deployment of embodied intelligence. In particular, basic manipulation tasks, such as point-to-point reaching and target grasping with robotic manipulators, impose stringent requirements on control accuracy, real-time response, and energy efficiency for edge deployment. As a prevailing end-to-end control paradigm, reinforcement learning (RL) has demonstrated remarkable capabilities in autonomous learning and generalization for decision-making in continuous action spaces. Nevertheless, constrained by electronic computing architectures, it generally suffers from substantial computing overhead, high inference latency and limited energy efficiency, making low-power edge deployment challenging. Optical computing, benefiting from its high-speed parallelism, offers a promising route to break energy-efficiency bottlenecks; however, research on the integration of optical computing and RL remains at an early stage. In this work, we propose a Mach-Zehnder Interferometer (MZI)-based opto-electronic hybrid RL architecture. We adopted two algorithm structures: Deep Deterministic Policy Gradient combined with Hindsight Experience Replay (DDPG-HER) and Twin Delayed Deep Deterministic Policy Gradient combined with HER (TD3-HER). We implemented the key layer linear computation of the spiking neural network (SNN) Actor on an optical MZI chip for the reaching task of a Franka Emika Panda robotic manipulator. The architecture is evaluated in the PyBullet simulation environment. Experimental results demonstrate that the proposed architecture achieves an energy efficiency of 1.39 TOPS/W and an inference latency of 120 ps while maintaining control accuracy, which provide a feasible pathway toward energy-efficient and precise control for robots.

Address all correspondence to Shuiying Xiang, email: syxiang@xidian.edu.cn; Changjian Xie, email: xiechangjian@xidian.edu.cn.

## Introduction

Dexterous object manipulation is a fundamental human capability supported by a sophisticated sensorimotor system. Through precise perception of the surrounding environment, rapid and iterative action planning, and adaptive behaviors guided by task objectives, humans can efficiently perform various manipulation tasks such as precise positioning and compliant grasping in complex unstructured scenarios. The unique musculoskeletal structure of the human hand, together with its multimodal sensory feedback mechanisms, enables precise target localization while allowing flexible adaptation to variations in object shape and weight, as well as environmental disturbances. Such innate dexterity has long served as an important source of inspiration for the development of robot manipulation technologies. Translating human manipulation principles and behavioral characteristics into robotic manipulator control systems to construct intelligent control frameworks capable of autonomous perception, autonomous decision-making and real-time execution, has become a major research focus in fields including service robots, intelligent industrial manufacturing, and specialized robots [1-4].

Intelligent and precise robotic manipulation relies heavily on the coordination of three fundamental components: perception, decision-making, and control [5]. First, high-precision environmental perception enables the system to acquire real-time information about target object poses, the robot's own state, and surrounding obstacles through sensors such as cameras and depth sensors. The acquired high-dimensional sensory data are then accurately processed and interpreted to provide reliable inputs for subsequent policy decisions. Second, precise policy generation involves rapidly producing stable and accurate joint control commands based on real-time sensory information within a continuous action space, while balancing response speed and positioning accuracy. Third, task-oriented optimization dynamically adjusts execution policies according to specific manipulation objectives, enabling the system to adapt to changing task conditions while maximizing execution efficiency and robustness. These three components work together to establish a closed-loop control framework, which is essential for achieving precise robotic manipulation.

The development of intelligent robotic manipulation has witnessed the emergence of three major control paradigms, driving progressive improvements in manipulation capabilities from structured to complex and unstructured environments. The earliest approaches primarily relied on control based on kinematic and dynamic modeling. By establishing accurate mathematical models of robotic manipulators, these methods solve the inverse kinematics problem to determine the required joint configurations and generate corresponding control commands for the end effector [6-8]. Although such approaches can achieve high positioning accuracy in structured environments, their performance is highly sensitive to model inaccuracies. Variations in object properties and environmental disturbances can significantly compromise control robustness, limiting their applicability in complex and unstructured environments.

The second paradigm is imitation learning, which acquires manipulation skills by learning from human demonstration trajectories. It can be rapidly deployed for specific tasks without elaborate modeling [9-11]. It features a straightforward training pipeline and strong task adaptability, yet its generalization ability heavily depends on the coverage of demonstration data, and performance degradation readily occurs when confronting unseen scenarios. The third paradigm is reinforcement learning (RL), which enables agents to optimize control policies through repeated trial-and-error interactions with their environments, without requiring explicit analytical models or predefined demonstrations. RL has demonstrated strong autonomous learning and generalization capabilities in control tasks involving continuous action spaces [12-15], making it a major research direction in intelligent robotic manipulation.

Advances in RL algorithms have continuously driven improvements in intelligent robotic manipulation. In 2013, Mnih et al. introduced the Deep Q-Network (DQN), which combines deep neural networks with Q-learning. Leveraging experience replay and target networks, DQN enabled end-to-end learning from high-dimensional visual observations to discrete actions [16]. In 2014, to address the challenges of continuous action spaces, Silver et al. proposed the Deterministic Policy Gradient (DPG) framework, which allows a deterministic policy to directly map states to continuous actions [17]. Building upon this framework, Lillicrap et al. proposed the Deep Deterministic Policy Gradient (DDPG) algorithm in 2015, combining deterministic policy gradients with deep neural networks and an Actor-Critic architecture to enable continuous control. The use of experience replay and soft updates of target networks further promotes training stability [18]. In 2017, Schulman et al. proposed Proximal Policy Optimization (PPO). By clipping the objective function to constrain the policy update step size, it balances sample efficiency and training stability, and has become a widely adopted policy optimization approach for continuous control tasks [19]. In 2018, the Twin Delayed Deep Deterministic Policy Gradient (TD3) algorithm was further introduced to address the overestimation of Q values in DDPG. By incorporating clipped double Q-learning, delayed policy updates, and target policy smoothing, TD3 effectively reduces value estimation errors and improves training stability and policy robustness [20]. Also in 2018, Haarnoja et al. proposed the Soft Actor-Critic (SAC) algorithm, which adopts the maximum entropy reinforcement learning framework. By maximizing both the expected cumulative reward and policy entropy, SAC encourages effective exploration and enables the learning of diverse stochastic policies, thereby improving adaptability to complex environments [21]. These advances have established a solid theoretical and methodological foundation for precise and autonomous robotic manipulation.

Meanwhile, recent advances in photonic neuromorphic computing have opened up new possibilities for overcoming the limitations of conventional electronic computing architectures [22-37]. Among these, optical neural networks (ONNs) based on the Mach-Zehnder Interferometer (MZI) exploit optical interference to perform matrix multiplication efficiently. Benefiting from inherent advantages including low latency, high parallelism, and energy efficiency, they have emerged as a promising approach to addressing computational bottlenecks in the post-Moore era [22-30]. In 2017, Shen et al. reported a programmable ONN based on MZI arrays and verified its potential for speech recognition [38]. In 2021, Xu et al. proposed an opto-electronic hybrid convolutional neural network (CNN) architecture that employs MZI arrays to perform convolution related computations optically, achieving a 1.73-fold improvement in energy efficiency over conventional electronic implementations [39]. In 2022, Wang et al. utilized soliton microcombs and wavelength division multiplexing (WDM) to experimentally demonstrate the feasibility of parallel computing using high-dimensional ONNs based on MZI arrays. Their work provided new architectural insights for the development of high-capacity on-chip optical AI accelerators [40].

Although the integration of RL and optical computing has demonstrated application potential, this research area remains at an early stage of development [41-43]. In 2023, Qiu et al. proposed a meta-surface based diffractive neural network and demonstrated the feasibility of using diffractive networks for RL-based decision making in simple game playing tasks, such as Tic-Tac-Toe [44]. However, their approach has not yet been extended to continuous action control of robotic manipulators. In 2024, Krauss et al. combined embedded optical waveguide sensor with a nonlinear autoregressive with exogenous inputs (NARX) neural network for state estimation in soft robotic fingers, which effectively reduced the end effector position estimation errors [45]. In 2025, Blackert et al. developed a soft robotic manipulator based on light responsive liquid crystal elastomers. They employed a spatial light modulator to generate an array of multiple light beams for shape control and used a CNN to infer the optical excitation parameters required to achieve target deformations [46]. Most existing studies focus on applying optical technologies to robotic sensing and actuation. By contrast, research on incorporating optical computing into RL decision-making networks for real-time robotic manipulator control remains limited. Consequently, significant challenges remain in integrating optical computing with RL-based control architectures.

In this work, we propose an MZI-based opto-electronic hybrid RL architecture for the point-to-point reaching task of a 7-degree-of-freedom Franka Emika Panda robotic manipulator. Two RL algorithms, DDPG with Hindsight Experience Replay (DDPG-HER) and TD3-HER, are employed. The Actor network adopts a spiking neural network (SNN) structure, and the time step is set to T = 4 to match the state update rhythm of the robotic manipulator. The MZI optical units are used to implement the linear transformation in the second layer of the Actor network. The Stochastic Parallel Gradient Descent (SPGD) algorithm is employed for offline training and precise phase configuration of MZI weights. During forward propagation, the input data undergoes linear transformation in the optical domain through the MZI array, exploiting the high speed and low power consumption of optical computing. The Critic network retains a conventional artificial neural network (ANN) architecture to maintain accurate value estimation and facilitate stable training convergence. A robotic reaching task is established in the PyBullet simulation environment to systematically evaluate the feasibility and performance of the proposed opto-electronic hybrid architecture. This work aims to provide a technical reference for the development of high speed, low power opto-electronic hybrid controllers for intelligent robots and to promote the practical application of optical computing in robotic control.

## Experimental Setup and Methods

### Opto-Electronic Hybrid Reinforcement Learning Architecture

DDPG is a classic Actor-Critic RL algorithm for continuous action space tasks [18]. The DDPG-based opto-electronic hybrid architecture constructed in this work is illustrated in Fig. 1(b). It consists of an Actor network for policy generation and a Critic network for value evaluation. The Actor network takes environmental observations including the current joint states of the robotic manipulator, the end-effector pose, and the target position as inputs, and outputs deterministic joint control actions to drive the manipulator toward the target. The Critic network takes state-action pairs as inputs and estimates the corresponding Q values. During training, the Critic network is updated by minimizing the temporal difference error, while the Actor network is optimized through policy gradients to maximize the Q value estimated by the Critic network. TD3 is an improved algorithm developed to address the overestimation of Q values in DDPG [20], as shown in Fig. 1(c). Building upon DDPG, TD3 incorporates three key mechanisms: clipped double Q learning, delayed policy update, and target policy smoothing. The clipped double Q-learning mechanism employs two independently trained Critic networks and uses the smaller of their target Q value estimates to effectively mitigate overestimation bias. Delayed policy updates reduce the update frequency of the Actor network relative to that of the Critic networks, thereby reducing the impact of value estimation errors on policy optimization. Target policy smoothing adds clipped noise to the target actions, reducing the sensitivity of value estimation to small variations in target actions and improving the robustness of policy learning.

The point-to-point reaching task of a robotic manipulator is a typical sparse reward problem. A positive reward is obtained only when the end effector successfully reaches the target region, whereas penalties are assigned to the other steps. This sparse reward structure makes it difficult for the agent to obtain informative learning signals and can consequently slow down policy convergence. To address this issue, Hindsight Experience Replay (HER) is incorporated into both the DDPG and TD3 frameworks [47]. The detailed algorithm procedure is illustrated in Table 1.

**Table 1 Hindsight Experience Replay (HER)**

---

**Algorithm Process**

---

**Given:**

- an off-policy RL algorithm $\mathbb{A}$,
- a strategy $\mathbb{S}$ for sampling goals for replay,
- a reward function $\boldsymbol{r} : \boldsymbol{\mathcal{S}} \times \boldsymbol{\mathcal{A}} \times \boldsymbol{\mathcal{G}} \rightarrow \boldsymbol{\mathcal{R}}$.

Initialize neural networks      Initialize replay buffer $\boldsymbol{R}$

**for** episode = 1, $M$ **do**

    Sample a goal $g$ and an initial state $\boldsymbol{s_0}$

    **for** $t = 0, T - 1$ **do**

        Sample an action at using the behavioral policy from $\mathbb{A}$: $\boldsymbol{a_t} \leftarrow \boldsymbol{\pi_b}(\boldsymbol{s_t}||\boldsymbol{g})$

        Execute the action at and observe a new state $\boldsymbol{s_{t+1}}$

    **end for**

    **for** $t = 0, T - 1$ **do**

        $\boldsymbol{r_t} := \boldsymbol{r}(\boldsymbol{s_t}, \boldsymbol{a_t}, \boldsymbol{g})$

        Store the transition $(\boldsymbol{s_t}||\boldsymbol{g}, \boldsymbol{a_t}, \boldsymbol{r_t}, \boldsymbol{s_{t+1}}||\boldsymbol{g})$ in $\boldsymbol{R}$

        Sample a set of additional goals for replay $\boldsymbol{G} := \mathbb{S}$(**current episode**)

        **for** $\boldsymbol{g'} \in \boldsymbol{G}$ **do**

            $\boldsymbol{r'} := \boldsymbol{r}(\boldsymbol{s_t}, \boldsymbol{a_t}, \boldsymbol{g'})$

            Store the transition $(\boldsymbol{s_t}||\boldsymbol{g'}, \boldsymbol{a_t}, \boldsymbol{r'}, \boldsymbol{s_{t+1}}||\boldsymbol{g'})$ in $\boldsymbol{R}$

        **end for**

    **end for**

    **for** $t = 1, N$ **do**

        Sample a minibatch $\boldsymbol{B}$ from the replay buffer $\boldsymbol{R}$

        Perform one step of optimization using $\mathbb{A}$ and minibatch $\boldsymbol{B}$

    **end for**

**end for**

---

HER improves learning efficiency through goal relabeling, transforming experiences from trajectories that fail to achieve the original goal into useful training experiences with respect to alternative goals. During experience replay, additional goals are sampled from the achieved states in the collected trajectory. Transitions that failed to reach the original goal are then relabeled with these alternative goals, and their corresponding rewards are recalculated according to the relabeled goals. In this way, experiences that fail to achieve the original goal can be transformed into successful transitions under alternative goals, providing additional learning signals for the agent. The relabeled experiences are stored together with the original experiences in the experience replay buffer and used for subsequent off-policy network updates. Without requiring modifications to the original reward function, HER improves sample efficiency by extracting useful information from unsuccessful attempts. This mechanism can accelerate policy convergence and enhance the adaptability of the learned policy to varying target positions and environmental disturbances, making it well suited to sparse reward robotic reaching tasks.

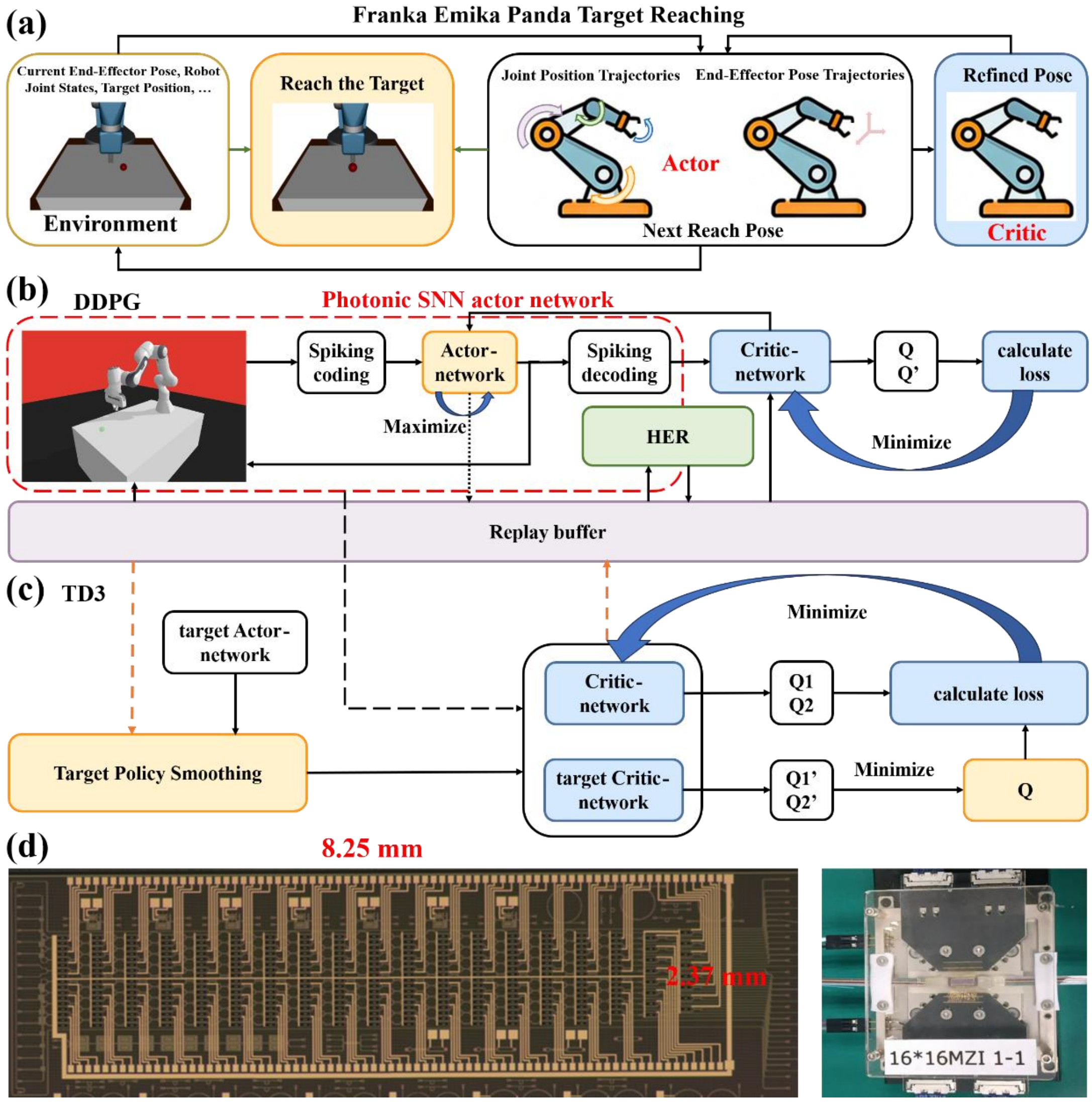

Figure 1. Overall system architecture. (a) Robotic manipulator control workflow; (b) DDPG algorithm architecture; (c) TD3 algorithm architecture; (d) Photographs of the chip before and after packaging.

To accelerate the linear computations in the Actor network using optical hardware, we designed and fabricated a 16×16 MZI photonic neuromorphic computing chip. Its microscopic structure and packaged prototype are shown in Fig. 1(d). The chip is fabricated on a silicon-on-insulator (SOI) platform, with an overall dimension of 2.37 mm×8.25 mm and a footprint of only 19.55 $mm^2$. It integrates 152 thermo-optic phase shifters based on titanium nitride (TiN) heaters, and supports parallel optical computation using real-valued matrices of size 16×16, with a matrix-vector multiplication (MVM) throughput of 2.5 TOPS. Based on the computational capacity of this 16×16 MZI chip, we customized the Actor network architecture to match the available optical hardware resources. As the policy generation unit, the Actor network adopts an SNN architecture, as illustrated in Fig. 2(a). Its dimensions are specifically configured to match the 16×16 computational capability of the MZI chip shown in Fig. 2(b). The network takes 12-dimensional state information as input, comprising the 6-dimensional manipulator state and velocity information, the 3-dimensional current end-effector position, and the 3-dimensional target position.

These inputs are converted into discrete spike sequences by a spike encoding layer using four discrete time steps (T = 4) and are then fed into the SNN hidden layer for nonlinear feature extraction. The 16× 16 linear multiplication operation in the second layer of the Actor network is implemented by the MZI optical array, replacing the conventional electronic linear layer. After offline calibration of MZI weights using the SPGD algorithm, the input spike sequences undergo parallel linear transformation in the optical domain through the MZI array during inference. The resulting outputs are converted by a spike decoding layer into 7-dimensional continuous joint control commands, which drive the robotic manipulator to perform the point-to-point reaching task. As the value evaluation unit, the Critic network retains a conventional fully connected artificial neural network (ANN) architecture and operates entirely in the electronic domain. It is not involved in optical hardware training and only guides the optimization of the Actor network through gradient information, thereby guaranteeing stability and accuracy throughout the training process.

**Environment Setup**

PyBullet [48] is an open source, high performance physics simulation engine for robotic learning, RL and autonomous control research. Based on rigid body dynamics and constraint solving algorithms, it supports multibody dynamics simulation, collision detection, friction modeling, and sensor noise simulation. It has been widely used in embodied intelligence research, including robotic manipulator control, dexterous manipulation, and mobile robot navigation. PyBullet provides an efficient framework for constructing simulation environments that reproduce the dynamic behavior of physical robots. For a 7-degree-of-freedom Franka Emika Panda robotic manipulator, it can simulate physical properties and dynamic effects such as joint rotations, link inertia, gravity, end-effector contact forces, joint friction, and actuator delays. It also supports customized sensor models and environmental disturbances, providing a controllable simulation platform for training and evaluating RL policies.

Based on PyBullet, we constructed a simulation environment for the point to point reaching task of the Franka Emika Panda robotic manipulator. The initial configuration of the manipulator is randomly generated within its safe operating workspace, while target positions are uniformly sampled within the 3-dimensional reachable workspace of the end effector. The workspace is defined by the following coordinate ranges: $x \in [0.2, 0.8]$ m, $y \in [-0.4, 0.4]$ m, $z \in [0.05, 0.8]$ m. To improve the robustness of the learned policy and its potential for transfer to real world scenarios, Gaussian random noise is injected into the simulation to model visual localization errors, joint angle measurement noise and actuator output disturbances. This allows the agent to learn under nonideal observation and action execution conditions. In addition, PyBullet provides real-time state feedback, enabling efficient access to the manipulator's joint positions, joint velocities, end effector pose, and interaction information with the surrounding environment. These observations provide stable and frequent state inputs for the RL algorithms, supporting efficient and reliable policy training.

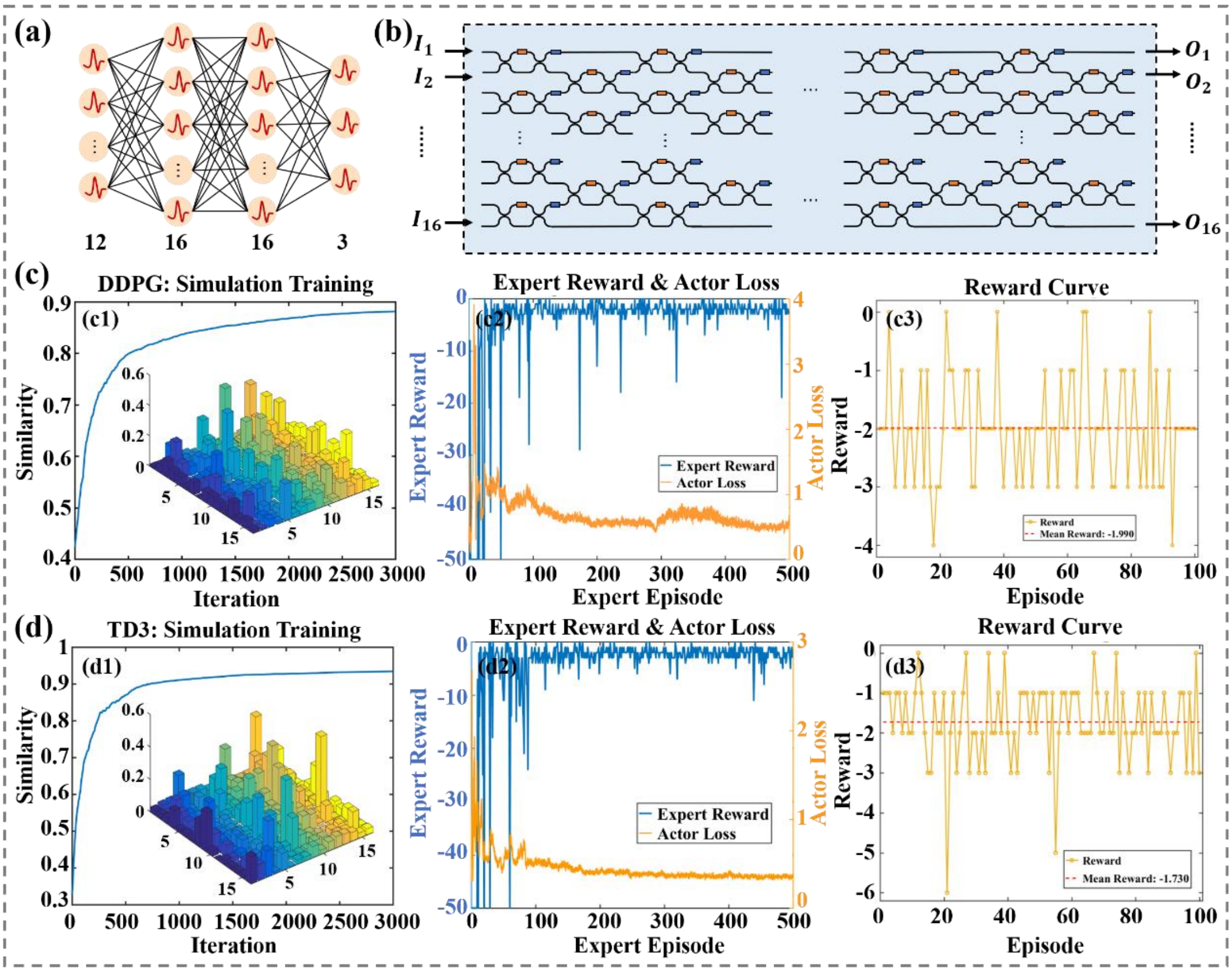

Figure 2. Simulation experiments. (a) Actor network dimensions; (b) MZI architecture; (c) Simulation results obtained using DDPG-HER; (c1) Simulation training curves and weights; (c2) Training reward and Actor loss curves; (c3) Test rewards; (d) Simulation results obtained using TD3-HER; (d1) Simulation training curves and weights; (d2) Training reward and Actor loss curves; (d3) Test rewards.

**Simulation Training**

After constructing the PyBullet simulation environment, we first performed pure software pre-training using the DDPG-HER and TD3-HER algorithms in the robotic manipulator reaching task environment to obtain converged network weights. The output matrix of the critical linear layer in the Actor network is taken as the target weight for optical implementation. Subsequently, an MZI array model was established in MATLAB, and the SPGD algorithm was employed to optimize the MZI phase settings. Cosine similarity was used as the optimization objective to progressively align the optical weights with the reference weight matrix, thereby configuring the desired optical computation. The simulation results show that the cosine similarity between the optical weights and the target weights approached 0.80 after 500 and 250 optimization iterations for DDPG-HER and TD3-HER, respectively. After 3000 iterations, the corresponding cosine similarity values reached 0.881 and 0.934, respectively, as shown in Fig. 2(c1) and Fig. 2(d1). After the weight configuration simulation, software fine-tuning was carried out within the algorithm to adapt the overall architecture to the characteristics of optical computation. This process yielded stable training reward curves for the hybrid architecture, as shown in

Fig. 2(c2) and Fig. 2(d2).

After policy convergence, the trained opto-electronic hybrid models were evaluated through 100 independent reaching task trials in the simulation environment, where the average reward and end-effector positioning accuracy are statistically evaluated. The test results are presented in Fig. 2(c3) and Fig. 2(d3). Over the 100 test trials, DDPG-HER achieved an average reward of -1.99 and an average end effector to target distance of 0.0326 m. TD3-HER achieved an average reward of -1.73 and an average distance of 0.0321 m. In this work, the success criterion was defined as an end effector to target distance of less than 0.05 m. Both algorithms satisfied this criterion in all 100 test trials, corresponding to an observed task success rate of 100%. These results demonstrate the feasibility of the proposed opto-electronic hybrid RL architecture and its ability to achieve accurate robotic reaching control in the simulation environment.

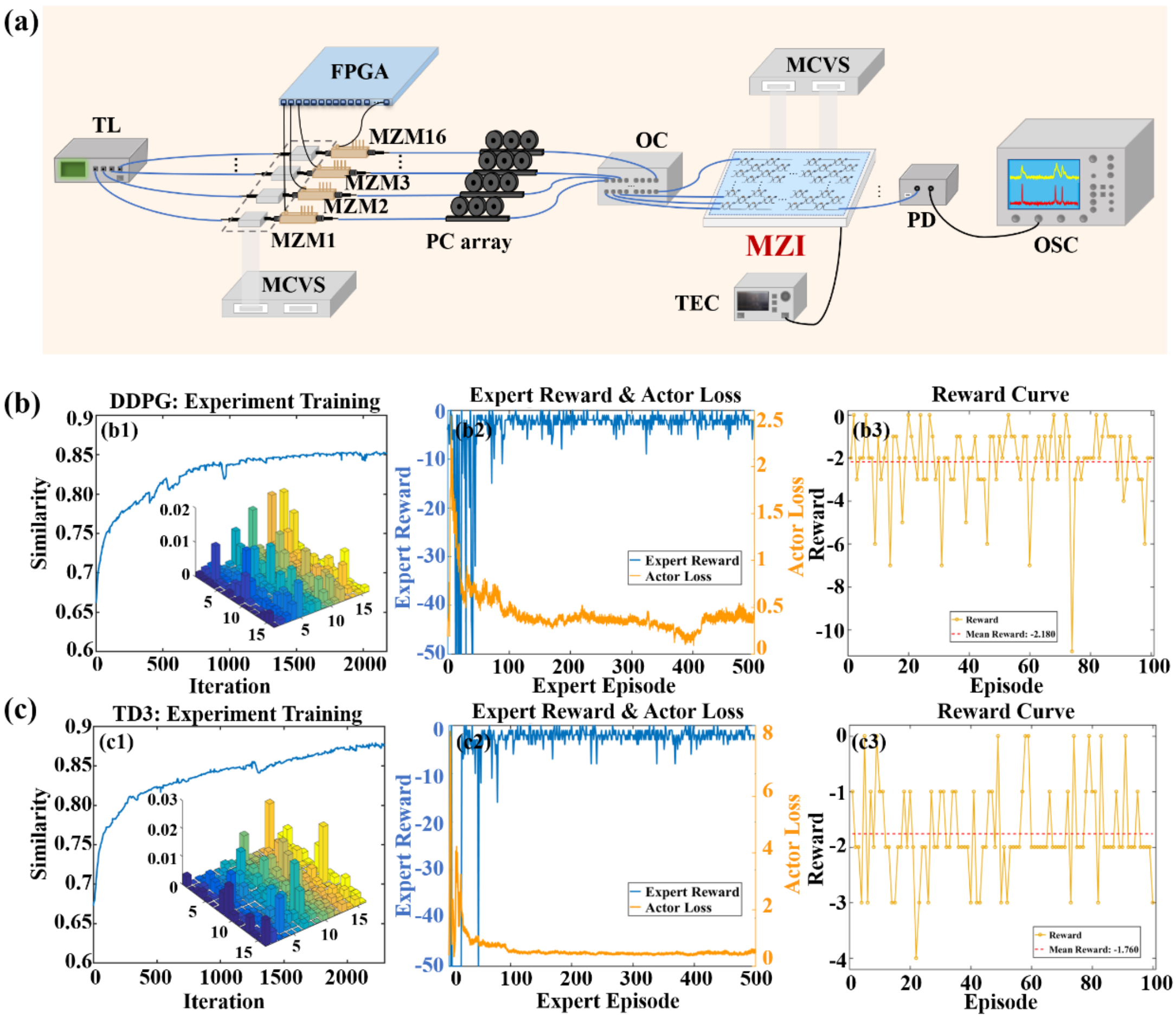

Figure 3. Hardware experiments setup and results. (a) Silicon-based MZI experimental platform; (b) Experimental results obtained using DDPG-HER; (b1) Experimental training curves and weights; (b2) Training reward and Actor loss curves; (b3) Test rewards; (c) Experimental results obtained using TD3-HER; (c1) Experimental training curves and weights; (c2) Training reward and Actor loss curves; (c3) Test rewards.

**Experimental Results and Analysis**

Following the simulation validation, we further conducted hardware experiments using the MZI

experimental platform shown in Fig. 3(a). The platform primarily consists of a tunable laser (TL), Mach-Zehnder modulator (MZM), polarization controller array (PC array), optical coupler (OC), 16×16 MZI optical computing chip, photodetector (PD), oscilloscope (OSC), and FPGA control unit. A thermoelectric cooler (TEC) is used to stabilize the chip operating temperature and suppress thermally induced phase drift, thereby reducing its impact on optical computing accuracy. A multi-channel voltage source supplies voltages to the phase shifters of the MZI array. The FPGA drives the MZM to load signals onto the optical carrier. The PC array is adjusted to establish the appropriate polarization states for optical transmission through the chip. The optical signals pass through the MZI array to complete linear computation. Finally, PDs convert optical signals into electrical signals, and the oscilloscope acquires the output optical intensity information.

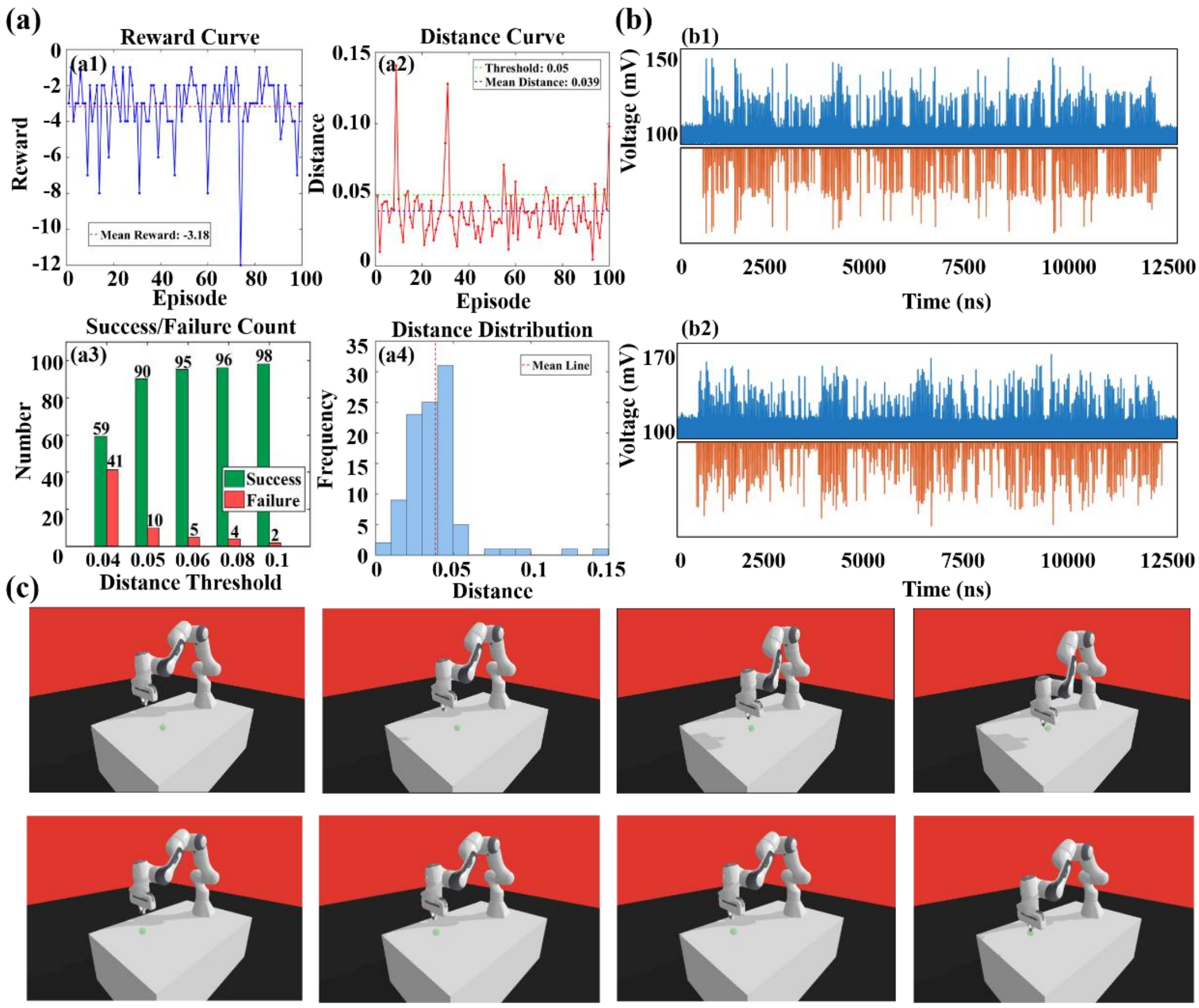

Figure 4. DDPG-HER experimental results. (a) Inference results obtained using DDPG-HER; (a1) Reward curve; (a2) End effector to target distance curve; (a3) Numbers of successful and failed trials under different distance thresholds; (a4) Distance distribution; (b) Timing diagram of the experimental inference process; (c) Two sets of robotic manipulator trajectories.

The target weight matrix of the key linear layer within the Actor network was extracted from the pre-trained network. The phase settings of the MZI array were then calibrated on the hardware platform using the SPGD algorithm. By optimizing the cosine similarity between the hardware implemented matrix and the target weight matrix, the effective optical weight matrix was progressively adjusted to approximate the target matrix. As shown in Figs. 3(b1) and 3(c1), the SPGD optimization for DDPG-HER and TD3-HER converged after approximately 2000 iterations, with the cosine similarity reaching stable values of 0.852 and 0.876, respectively. These values are slightly lower than those obtained in simulation, mainly due to nonideal effects such as measurement noise and phase drift. The equivalent MZI matrices measured on the hardware platform were then imported into the simulation environment. One hundred independent point-to-point reaching tests of the robotic manipulator were performed, and the average reward, positioning accuracy and task success rate were statistically analyzed. The experimental results are illustrated in Fig. 3(b3) and 3(c3). DDPG-HER achieved an average reward of -2.18 and an average distance of 0.0343 m to the target position. TD3-HER obtained an average reward of -1.76 and an average distance of 0.0332 m. Both algorithms achieved a task success rate of 100.0%.

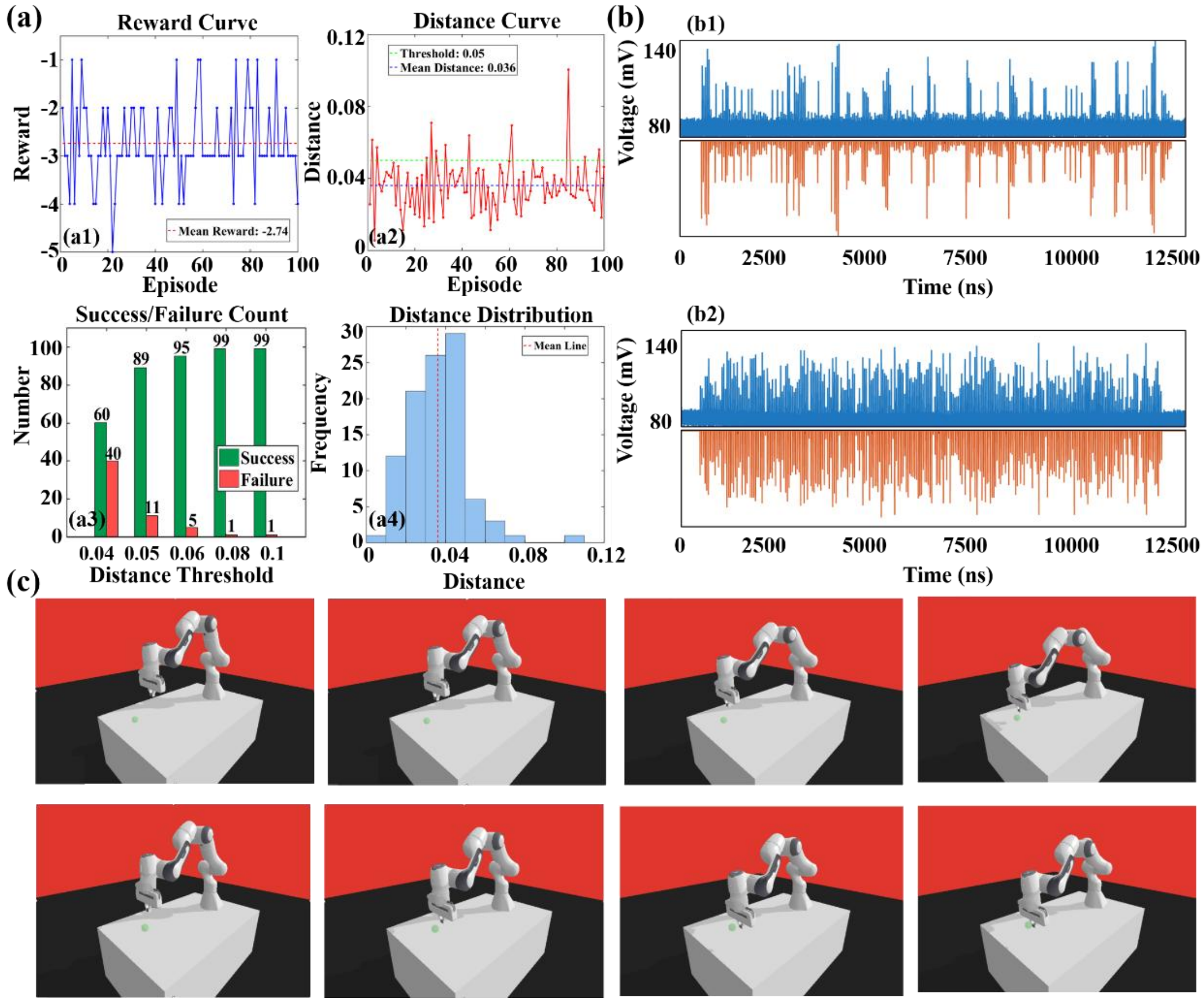

Figure 5. TD3-HER experimental results. (a) Inference results obtained using TD3-HER; (a1) Reward curve; (a2) End effector to target distance curve; (a3) Numbers of successful and failed trials under different distance thresholds; (a4) Distance distribution; (b) Timing diagram of experimental inference; (c) Two sets of robotic manipulator trajectories.

After completing the phase calibration of the chip, the MZI photonic chip was adopted to perform optical-domain forward computation for the linear layer of the Actor network, and further inference experiments were carried out. The experimental results are presented in Fig. 4 and Fig. 5. The average reward of DDPG-HER was -3.18, and the average distance between the end-effector and the target position was 0.0386 m. For TD3-HER, the average reward was -2.74 with an average end-effector distance of 0.0360 m. Statistics of task success rates under different reaching distance thresholds reveal that when the distance threshold ranges from 0.05 m to 0.1 m, the average task success rates of both DDPG-HER and TD3-HER exceed 90%, and TD3-HER outperforms DDPG-HER overall.

Figures 4(b1) and 4(b2) show the hardware output timing diagrams of two randomly selected channels of the DDPG-HER model, corresponding to channel 1 at T = 2 and channel 10 at T = 3, respectively. Similarly, Figs. 5(b1) and 5(b2) present the hardware output timing diagrams of two randomly selected channels of the TD3-HER model, corresponding to channel 7 at T = 2 and channel 5 at T = 4, respectively. In all four cases, the measured hardware outputs match the theoretical target outputs well, with only minor deviations. Figures 4(c) and 5(c) illustrate two sets of robotic manipulator trajectories obtained using DDPG-HER and TD3-HER, respectively. In both cases, the manipulator end-effector gradually approaches the target position along the planned trajectories, demonstrating that both algorithms can effectively guide the manipulator toward the target during hardware inference.

For the optical linear computation stage, the MZI array exhibits an optical propagation latency as low as 120 ps and a computational energy efficiency of up to 1.39 TOPS/W. Compared with conventional electronic computing architectures, the optical computing unit achieves remarkable latency reduction and substantial improvement in energy efficiency. While guaranteeing control accuracy, it attains dual breakthroughs in real-time performance and power consumption, which well meets the deployment requirements of edge robots for low power consumption and high real-time responsiveness.

## Conclusion

This work proposes an MZI-based opto-electronic hybrid RL architecture for the point-to-point reaching task of the Franka Emika Panda robotic manipulator, incorporating DDPG-HER and TD3-HER. The MZI array is employed to implement the key linear computation layer of the SNN-based Actor network, and the SPGD algorithm is employed for phase configuration of optical weights. The feasibility and performance of the architecture are evaluated through PyBullet simulations and hardware inference experiments. Experimental results demonstrate that after introducing the MZI photonic hardware, the system control performance degrades to a certain extent compared with pure software simulation due to nonideal factors such as chip noise and thermal phase drift. Nevertheless, both algorithms can still accomplish the robotic manipulator point-to-point reaching task. During hardware inference, DDPG-HER achieves an average reward of -3.18 and an average distance of 0.0386 m between the end effector and the target, while TD3-HER obtains an average reward of -2.74 and an average distance of 0.0360 m. Within the distance threshold range of 0.05-0.1 m, the average task success rates of DDPG-HER and TD3-HER exceed 90%. The MZI array achieves an optical propagation latency as low as 120 ps and a computational energy efficiency of 1.39 TOPS/W. While guaranteeing control accuracy, it significantly improves real-time performance, providing a feasible approach toward low power and high-speed robotic manipulation.

## Discussion

Focusing on the application of the opto-electronic hybrid RL architecture in robotic manipulator control, this work carries out full-process experiments including simulation calibration, MZI hardware phase calibration and MZI hardware inference. The performance differences between DDPG-HER and TD3-HER are quantitatively evaluated using two metrics: average reward and end-effector positioning error. Overall, TD3-HER exhibits better control performance than DDPG-HER across all evaluated stages, including software simulation, simulation evaluation using hardware-calibrated MZI matrices, and hardware inference. In the simulation environment, TD3-HER achieves an average reward of -1.73, compared with -1.99 for DDPG-HER, and an average positioning error of 0.0321 m, compared with 0.0326 m for DDPG-HER. These results indicate that TD3-HER achieves slightly better control performance than DDPG-HER under software simulation conditions. When hardware-calibrated MZI weight matrices are imported into the simulation environment for evaluation, both algorithms exhibit only slight performance degradation, suggesting that the MZI implementation preserves much of the functionality of the target linear transformations. During hardware inference, the overall system performance degrades due to hardware noise, phase drift, and non-ideal device characteristics. Nevertheless, TD3-HER achieves an average reward of -2.74 and an average positioning error of 0.0360 m, outperforming DDPG-HER, which achieves an average reward of -3.18 and an average positioning error of 0.0386 m. These results demonstrate that, despite the performance degradation introduced by hardware implementation, both algorithms maintain effective reaching performance, with TD3-HER consistently achieving better results under the tested conditions. Overall, both algorithms maintain centimeter-level reaching accuracy during hardware inference.

**Table 2 Experimental Performance Comparison**

| RL Algorithm | Performance Metrics | Simulation | MZI Training | MZI Experiment |
|---|---|---|---|---|
| **DDPG** | Mean Reward | -1.99 | -2.18 | -3.18 |
| | Mean Distance (m) | 0.0326 | 0.0343 | 0.0386 |
| **TD3** | Mean Reward | -1.73 | -1.76 | -2.74 |
| | Mean Distance (m) | 0.0321 | 0.0332 | 0.0360 |

The comparison indicates that the performance degradation introduced by the photonic neuromorphic platform is limited, while the optical linear computation stage offers low latency and high computational energy efficiency. The proposed opto-electronic hybrid RL architecture demonstrates a promising balance between control accuracy and computational efficiency, providing a potential implementation approach for intelligent decision-making systems requiring low-latency and energy-efficient processing.

## Data Availability

The data that support the findings of this study are available from the corresponding author upon reasonable request.

## Research Funding

This work was supported by the National Key Research and Development Program of China (2021YFB2801900); National Natural Science Foundation of China (No. 62535015, 62575231); The China Postdoctora Science Foundation (Certificate Number: 2024M752526, 2024M762537); The National Science Foundation for Young Scientists of China (62404173); Xidian University Specially Funded Project for Interdisciplinary Exploration (TZJH2024009).

## Competing Financial Interests

The authors declare no competing financial interest.